\documentclass[10pt,twocolumn,letterpaper]{article}

\usepackage{wacv}
\usepackage{times}
\usepackage{epsfig}
\usepackage{graphicx}
\usepackage{amsmath}
\usepackage{amssymb}
\usepackage[mode=buildnew]{standalone}
\usepackage[font=small]{subfig}
\usepackage{booktabs}
\usepackage{multirow}
\usepackage{siunitx}
\usepackage[english]{babel}
\usepackage{xcolor}
\usepackage{blindtext}
\usepackage{hyperref}
\usepackage{mathtools}
\usepackage{tabularx}
\usepackage{stackengine}
\usepackage{cases}

\newcommand{\myblindtext}[1]{}%\textcolor{white}{\blindtext[#1]}}
\DeclarePairedDelimiterX{\norm}[1]{\lVert}{\rVert}{#1}

\wacvfinalcopy % *** Uncomment this line for the final submission

\def\wacvPaperID{542} % *** Enter the wacv Paper ID here

\ifwacvfinal\fi
\begin{document}

%%%%%%%%% TITLE
\title{Improving Performance of Semantic Segmentation CycleGANs\\by Noise Injection into the Latent Segmentation Space}

% Authors at the same institution
%\author{First Author \hspace{2cm} Second Author \\
% Authors at different institutions
\author{Jonas Löhdefink, Tim Fingscheidt\\
Technische Universität Braunschweig\\
{\tt\small \{j.loehdefink, t.fingscheidt\}@tu-bs.de}
}

\maketitle
\ifwacvfinal\thispagestyle{empty}\fi

%%%%%%%%% ABSTRACT
\begin{abstract}
	In recent years, semantic segmentation has taken benefit from various works in computer vision.
	Inspired by the very versatile CycleGAN architecture, we combine semantic segmentation with the concept of cycle consistency to enable a multitask training protocol.
	However, learning is largely prevented by the so-called steganography effect, which expresses itself as watermarks in the latent segmentation domain, making image reconstruction a too easy task.
	To combat this, we propose a noise injection, based either on quantization noise or on Gaussian noise addition to avoid this disadvantageous information flow in the cycle architecture.
	We find that noise injection significantly reduces the generation of watermarks and thus allows the recognition of highly relevant classes such as ``traffic signs'', which are hardly detected by the ERFNet baseline.
	We report mIoU and PSNR results on the Cityscapes dataset for semantic segmentation and image reconstruction, respectively.
	The proposed methodology allows to achieve an mIoU improvement on the Cityscapes validation set of \SI{5.7}{\percent} absolute over the same CycleGAN without noise injection, and still an absolute \SI{4.9}{\percent} over the ERFNet non-cyclic baseline.
\end{abstract}
\vspace*{-0.25cm}

%%%%%%%%% BODY TEXT

\section{Introduction}

% Kontexteinführung, Relevanz
Multi-task learning helps in optimizing neural networks.
So it is common knowledge that the extension of supervised semantic segmentation training by an image reconstruction step as additionally-learned task is improving performance~\cite{mondal2019revisiting, Schlemper2017}.
The advantage of image reconstruction is that no additional labels are required for the training of an image $\rightarrow$ segmentation $\rightarrow$ image cycle, since the input image serves as training target~\cite{Zhu2017}.
Speaking in terms of an autoencoder, the segmentation predictions can be considered as latent space data in the so-called latent segmentation space.
We conjecture that cycle consistency will help in learning the semantic segmentation function.

% CycleGAN Teaser
\begin{figure}[t!]
	\centering
%	\includestandalone{figures/teaser}
	\includegraphics{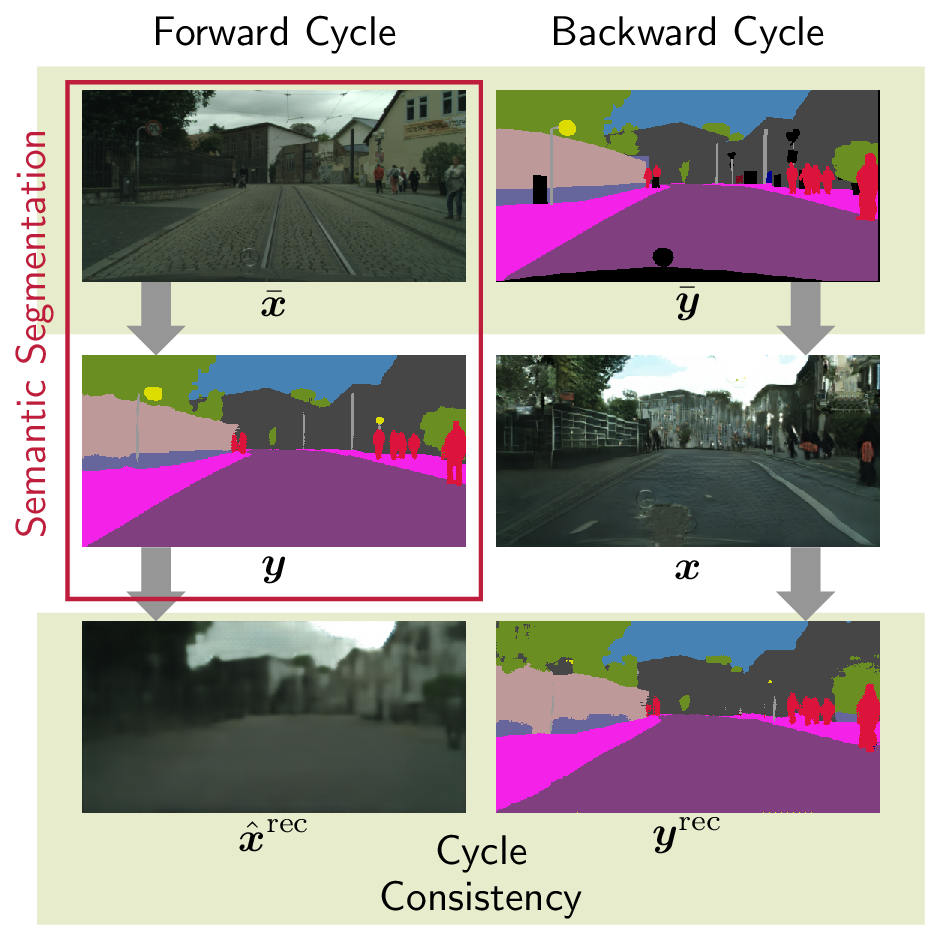}
	\caption
	{
		% Title
		\textbf{Examples} from the cycle-consistent segmentation framework.
		% Two Cycles
		The left side shows the \textbf{forward cycle}: real input image $\bar{\boldsymbol{x}}$, segmentation prediction $\boldsymbol{y}$ (to be compared to the respective ground truth $\bar{\boldsymbol{y}}$), and reconstruction of the input image $\hat{\boldsymbol{x}}^\textrm{rec}$.
		The right side shows the \textbf{backward cycle}: segmentation input $\bar{\boldsymbol{y}}$, synthetic image prediction $\boldsymbol{x}$, and reconstruction of the input segmentation $\boldsymbol{y}^\textrm{rec}$.
		% Teaser
		Cycle consistency facilitates learning of the actual semantic segmentation task and would easily allow the extension towards unpaired data.
	}
	\label{fig:intro}
	\vspace*{-0.5cm}
\end{figure}

% Figure
The left column of Figure~\ref{fig:intro} shows samples of the forward cycle (image $\rightarrow$ segmentation $\rightarrow$ image), containing the semantic segmentation operation we focus at in this work, while the right column shows the backward cycle (segmentation $\rightarrow$ image $\rightarrow$ segmentation).
Both cycles are learned jointly in the framework.
Though the architecture has a promising potential to reduce the need of labeled data by semi-supervised training setups~\cite{mondal2019revisiting}, in this work we concentrate on im\-pro\-ving the cycle-consistent segmentation framework, using only the Cityscapes dataset~\cite{Cordts2016}.

% Introduce watermark problem
The occurrence of steganography~\cite{Liu2019f} is a common issue in cycle-consistent architectures and can be imagined as a watermark---effectively invisible for observers---contained in the least-significant bits of the hidden parts of the latent representation, but bearing important information about the input image~\cite{Chu2017}.
The image-reconstructing function in a CycleGAN framework (segmentation $\rightarrow$ image step in the forward cycle) tends to rely largely on these watermarks for the reconstruction of the original image.
However, we want the networks to learn a good image representation instead of hiding side information in a signal, for much easier image reconstruction.
To overcome this problem, we prevent pro\-pa\-ga\-tion of watermarks into the image resynthesis network by injecting various types of noise into the forward cycle with the purpose of destroying such information.
We conjecture and will show that the resulting more powerful image resynthesis network will also provide benefits for our envisaged improvement of the semantic segmentation.

% Contributions
Our contributions comprise first, the adaptation of the CycleGAN framework~\cite{Zhu2017} to the architecture of the ERFNet~\cite{Romera2018}, second, the proposal of a noise injection into the latent segmentation space of the cycle-consistent segmentation architecture with various noise types, and third, analysis on the effects of the different noise injections on the segmentation and image reconstruction performance, measured by mIoU and PSNR, respectively.

% Structure of Paper
The remainder of this paper is structured as follows.
In Section~\ref{sec:rel_work}, an overview about related work can be found.
Afterwards, in Section~\ref{sec:methodology}, we will explain our cycle-consistent framework along with the noise injection.
Section~\ref{sec:eval} will present the experimental setup and results of the conducted investigations.
Finally, we conclude our work in Section~\ref{sec:conclusion}.

\section{Related Work}
\label{sec:rel_work}
This section provides an overview about important works in related research fields.

\textbf{Semantic Segmentation.}
\label{sec:rel_work_seg}%
Semantic segmentation is the dense (pixel-wise) prediction of probability distributions for an input image to be in one of the predefined semantic classes.
A comprehensive overview over the state-of-the-art in semantic segmentation is given in~\cite{minaee2020image}.
Most of the semantic segmentation networks are based upon fully convolutional networks (FCNs)~\cite{Long2015, Shelhamer2017} and have an encoder-decoder design~\cite{Noh2015}.
% Feature extractor and segmentation head
Many segmentation frameworks further propose to use a generic feature extractor~\cite{Chen2018, Chen2018a, Zhao2016a} with a subsequent segmentation head, performing the actual segmentation.
The segmentation head rebuilds the image resolution by means of bilinear interpolation~\cite{Chen2018, RotaBulo2018} or dilated convolutions~\cite{Chen2015} and uses techniques such as skip connections~\cite{Bilinski2018, Chen2018a}, spatial pyramid pooling~\cite{Chen2018, Chen2018a}, fractional residual units~\cite{Romera2018}, and depthwise separable convolution \cite{Chollet2017}.
Addressing real-time applicability, some architectures are explicitly designed to allow high computational efficiency~\cite{Baer2020, RotaBulo2018, Romera2018}.
Semantic segmentation is a key technology for many applications such as perception for autonomous driving~\cite{Baer2019, Bolte2019}.

We use the widely-employed ERFNet~\cite{Romera2018}, since it is an efficient topology, while still providing a decent segmentation performance on the Cityscapes dataset~\cite{Cordts2016}.

\textbf{Image-to-Image Translations.}
\label{sec:rel_work_cycle}%
% GAN
Generative adversarial networks (GANs)~\cite{Goodfellow2014} have become standard in computer vision and were extended to use conditional input~\cite{Mirza2014} as well as various loss functions~\cite{Arjovsky2017, Mao2016, Rubner2000}.
Image-to-image translation~\cite{Isola2016, Pizzati2020} is the conversion of images between two different domains, which is inversely related to domain adaptation~\cite{Bolte2019a, Loehdefink2020}, where the models try to be domain-agnostic.
While GANs perform well in this application, the CycleGAN~\cite{Zhu2017, mondal2019revisiting} is the most famous approach, enforcing cycle consistency.
Since the reconstruction loss penalizes large differences between original and reconstructed signal, the networks embed non-noticeable information into the latent space, which is used to reconstruct the input image with surprisingly high accuracy.
This effect is known from the field of cryptography under the term \textit{steganography}~\cite{Liu2019f}, which often occurs in cycle-consistent frameworks~\cite{Chu2017, Khan2020, mondal2019revisiting}, and is not restricted to images as input only~\cite{Ye2019b}.
Referring to such hidden information in the segmentation space, we will call it watermarks in this work.

Our CycleGAN framework for semantic segmentation builds upon Mondal et al.\ \cite{mondal2019revisiting}, but while they use a generic image-to-image network, we use the ERFNet~\cite{Romera2018} and concentrate on avoiding watermark effects to challenge image reconstruction and thereby to achieve better semantic segmentation performance.

% Architecture
\begin{figure*}[t!]
	\centering
	\includegraphics{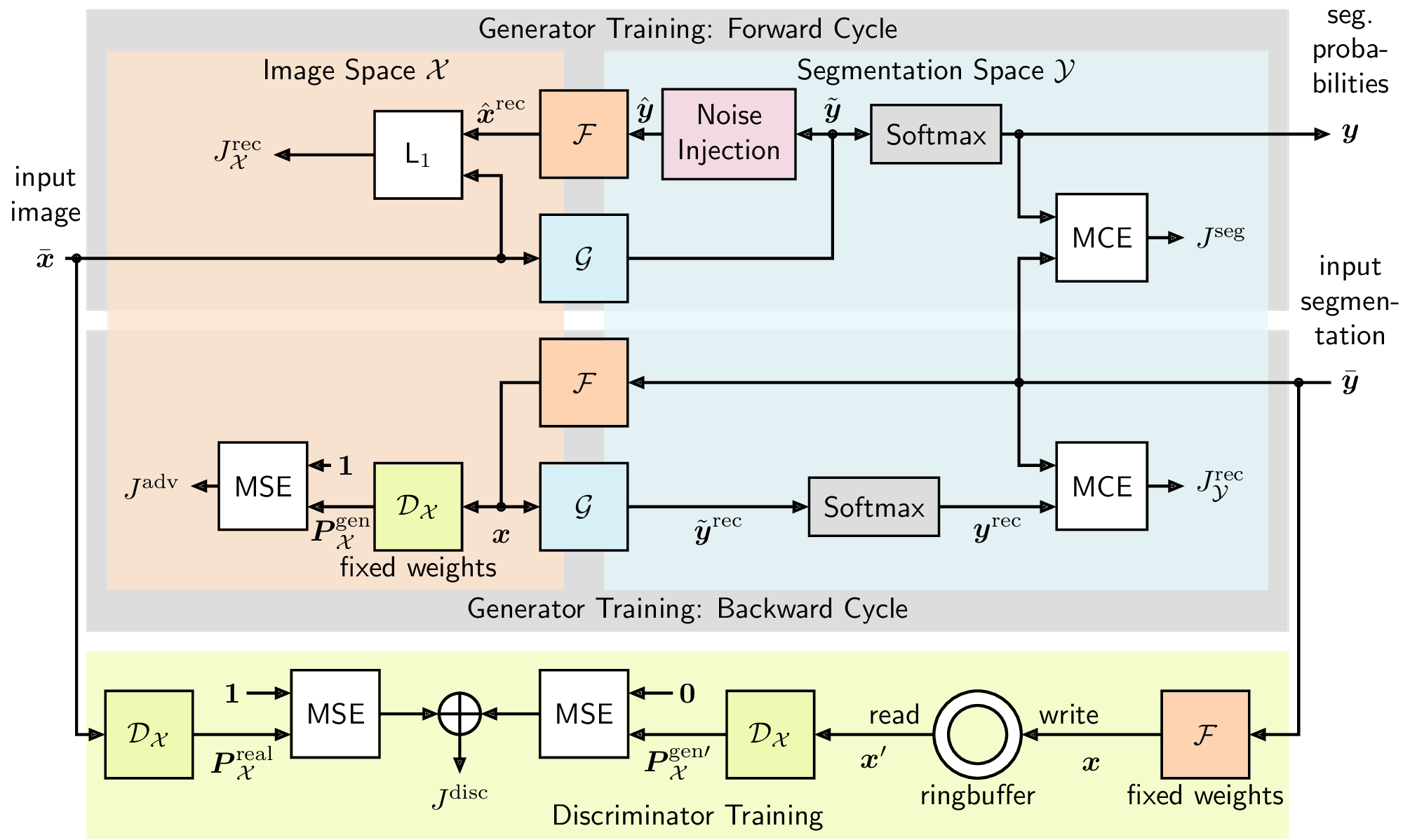}
	\caption{
		\textbf{Cycle-consistency-based architecture} for semantic segmentation divided into image space ($\mathcal{X}$) and segmentation space ($\mathcal{Y}$).
		Our proposed framework uses the (``ground truth'') images $\bar{\boldsymbol{x}}$ and the respective ground truth semantic segmentation labels $\bar{\boldsymbol{y}}$ to train a framework for semantic segmentation.
		The generators $\mathcal{G}$ and $\mathcal{F}$ are trained by the generator loss $J^\textrm{gen}$ \eqref{eq:gen_loss} to translate the inputs between $\mathcal{X}$ and $\mathcal{Y}$, where $\mathcal{G}$ constitutes the semantic segmentation network.
		The adversarial loss is implemented by a discriminator network $\mathcal{D}_\mathcal{X}$, being simultaneously trained with the framework (lower part of the figure), and assessing the quality of synthetic images generated by $\mathcal{F}$ by estimating a vector of probabilities $\boldsymbol{P}_\mathcal{X}^\textrm{gen}$ that the synthesized image $\boldsymbol{x}$ is real~\cite{Goodfellow2014}.
		We propose to inject noise into the segmentation predictions $\tilde{\boldsymbol{y}}$, before translating back into the image space (before softmax), to prevent the networks from propagating image-specific watermarks for a too-simple image reconstruction, leading to a higher segmentation performance.
	}
	\label{fig:cycle_gan}
%	\vspace*{-0.25cm}
\end{figure*}

\textbf{Image Restoration and Compression.}
\label{sec:rel_work_noise}%
% Image Restoration
Image restoration with neural networks is often times based on CNNs and describes the task of recovering an image from noise or from a blurred version~\cite{Tai2017, Zhang2017a}.
While being application-agnostic, these techniques are used, \eg, in medical applications to accelerate data acquisition by fast decoders~\cite{Schlemper2017, Zhang2017a}.
% Compression
With the use of machine learning, also image compression has experienced large improvements~\cite{Agustsson2018, Loehdefink2019, Theis2017}.
Most of its methods again are based on autoencoders~\cite{Agustsson2018, Rippel2017, Toderici2015}, consisting of CNNs and using transposed convolutions in the decoder~\cite{Loehdefink2020IV, zeiler2014visualizing, zeiler2010deconvolutional}.
Also here, adversarial loss functions take an important role~\cite{Agustsson2018, Rippel2017, Santurkar2017}.
% Quantization
Lossy compression methods rely on quantization, which can be divided into scalar~\cite{Agustsson2018, Yang_2019_CVPR, zha2005} and vector quantization~\cite{Gray1984, gersho2012, agustsson2017soft, loehdefink20ITSC}.
In vector quantization, groups of input values---instead of scalars---are assigned to one symbol~\cite{Gray1984, Gray1990}, thereby improving the compression performance.
Concerning evaluation, the human-perceived image quality is different from instrumental image assessment metrics such as PSNR~\cite{Zhang2012, Zhao2016}.
Hence there is research on better-suited loss functions~\cite{ma2017learning, talebi2018nima}, obtaining more pleasing results, \eg, based on the structural similarity (SSIM)~\cite{wang2004image, Zhao2016b}.

Inspired by quantization methods in image compression, we use uniform scalar quantizers with straight-through-estimation (STE)~\cite{bengio-arxiv13-condcomp}, or alternatively a Gaussian noise generator and adder as our proposed noise injection methods.

\section{Method}
\label{sec:methodology}
This section will introduce the proposed cycle-consistency-based framework for semantic segmentation that forms the basis of our investigations.
In addition, the noise-generating functions used in our framework to reduce the impact of watermarks will be presented.

\subsection{Cycle-Consistency-Based Architecture and Losses}
\label{sec:methodology_architecture}
Figure~\ref{fig:cycle_gan} shows our proposed cycle-consistency-based segmentation architecture.
First, the architecture is \mbox{addressed}, then the loss functions.

% Image Description
\textbf{Architecture.}
The framework is divided into the image space $\mathcal{X}$, where all signals have \mbox{$|\mathcal{C}| = C = 3$} color channels with \mbox{$\mathcal{C} = \{r, g, b\}$}, and the segmentation space $\mathcal{Y}$, where the signals have \mbox{$|\mathcal{S}| = S$} classes with \mbox{$\mathcal{S} = \{1,2,...,S\}$}~\cite{mondal2019revisiting}.
The pixel index \mbox{$i\in \mathcal{I} = \{1,2,..., H\!\cdot\!W\}$} with image height $H$ and width $W$ holds for all signals in the image and segmentation space.
The inputs are the (``ground truth'') images $\bar{\boldsymbol{x}} = (\bar{\boldsymbol{x}}_i)$ with the normalized (color) pixels $\bar{\boldsymbol{x}}_i\in [-1, 1]^C$ and the ground truth semantic segmentation labels $\bar{\boldsymbol{y}} = (\bar{\boldsymbol{y}}_i)$ with the one-hot segmentation pixels $\bar{\boldsymbol{y}}_i \in \{0,1\}^S$.
The two generators $\mathcal{G}$ and $\mathcal{F}$ translate the inputs from $\mathcal{X}$ to $\mathcal{Y}$, and vice versa, respectively.
Starting in the image space, $\mathcal{G}$ produces segmentation predictions $\tilde{\boldsymbol{y}} = (\tilde{\boldsymbol{y}}_i)$ with $\tilde{\boldsymbol{y}}_i \in \mathbb{R}^S$ from the ground truth images $\bar{\boldsymbol{x}}$, which are then subject to the softmax function to obtain the segmentation probabilities $\boldsymbol{y} = (\boldsymbol{y}_i)$ with $\boldsymbol{y}_i \in [0,1]^S$.
This can be considered as standard semantic segmentation path.
Additionally, in our proposed architecture, we inject noise into the segmentation predictions $\tilde{\boldsymbol{y}}$, obtaining the noisy segmentation predictions $\hat{\boldsymbol{y}} = (\hat{\boldsymbol{y}}_i)$ with $\hat{\boldsymbol{y}}_i \in \mathbb{R}^S$ (see Section~\ref{sec:methodology_noise}), and retranslate them via the second generator $\mathcal{F}$ back to $\mathcal{X}$, making up the image reconstructions $\hat{\boldsymbol{x}}^\textrm{rec} = (\hat{\boldsymbol{x}}^\textrm{rec}_i)$ with $\hat{\boldsymbol{x}}^\textrm{rec}_i \in \mathbb{R}^C$ and the forward cycle of the architecture.
The backward cycle, in contrast, starts in the segmentation space, generating image predictions $\boldsymbol{x} = (\boldsymbol{x}_i)$ with $\boldsymbol{x}_i \in \mathbb{R}^C$ from segmentation labels $\bar{\boldsymbol{y}}$.
A discriminator $\mathcal{D}_\mathcal{X}$, which is trained in parallel but has fixed weights during generator training, afterwards estimates a realism probability vector $\boldsymbol{P}^\textrm{gen}_\mathcal{X} \in [0, 1]^{H_\textrm{D} \times W_\textrm{D}}$, with height $H_\textrm{D}$ and width $W_\textrm{D}$ stemming from the specific downsampling architecture in the discriminator~\cite{Zhu2017}, typically delivering real/fake decisions for smaller image patches.
This vector represents the probability that the discriminator's input is real rather than generated, which is to be maximized by the generator $\mathcal{F}$.
Transforming $\boldsymbol{x}$ via $\mathcal{G}$ back to the segmentation space $\mathcal{Y}$ yields the segmentation reconstruction predictions $\tilde{\boldsymbol{y}}^\textrm{rec} = (\tilde{\boldsymbol{y}}^\textrm{rec}_i)$ with $\tilde{\boldsymbol{y}}^\textrm{rec}_i \in \mathbb{R}^S$, which can be converted to the segmentation reconstruction probabilities $\boldsymbol{y}^\textrm{rec} = (\boldsymbol{y}^\textrm{rec}_i)$ with $\boldsymbol{y}^\textrm{rec}_i \in [0,1]^S$ again.

% Losses (incl. Discriminator)
\textbf{Losses.}
The loss function used in \textit{generator training} consists of four contributions: First, the standard semantic segmentation loss~\cite{minaee2020image, Shelhamer2017}
\begin{equation}
J^\textrm{seg} = \operatorname{MCE}(\boldsymbol{y}, \bar{\boldsymbol{y}}) = -\frac{1}{H W} \sum_{i \in \mathcal{I}} \bar{\boldsymbol{y}}_i^\textrm{T} \log \boldsymbol{y}_i,
\end{equation}
with the image height $H$ and width $W$, the log taken element-wise, and $()^\textrm{T}$ being the transpose, computing the minimum cross entropy (MCE) between segmentation probabilities $\boldsymbol{y}$ and ground truth segmentation labels $\bar{\boldsymbol{y}}$, being the only loss requiring paired data.
Second, the adversarial loss~\cite{Mao2016} being the mean squared error (MSE)
\begin{equation}
J^\textrm{adv} = \operatorname{MSE}(\boldsymbol{P}^\textrm{gen}_\mathcal{X}, \boldsymbol{1}) = \frac{1}{H_\textrm{D} W_\textrm{D}} \norm{\boldsymbol{P}^\textrm{gen}_\mathcal{X} - \boldsymbol{1}}_2^2,
\end{equation}
between the realism probability vector $\boldsymbol{P}^\textrm{gen}_\mathcal{X} = \mathcal{D}_\mathcal{X}(\boldsymbol{x})$ and the vector of ones $\boldsymbol{1}$, both with height $H_\textrm{D}$ and width $W_\textrm{D}$.
Here, the simultaneously trained discriminator $\mathcal{D}_\mathcal{X}$ is used, which classifies its input into real or fake, and operates on the image predictions $\boldsymbol{x}$.
Third and fourth, the reconstruction losses (see Fig.~\ref{fig:cycle_gan}) \cite{mondal2019revisiting}
\begin{equation}
J_\mathcal{X}^\textrm{rec} = \operatorname{L}_1(\hat{\boldsymbol{x}}^\textrm{rec}, \bar{\boldsymbol{x}}) = \frac{1}{H W} \norm{\hat{\boldsymbol{x}}^\textrm{rec} - \bar{\boldsymbol{x}}}_1
\end{equation}
being the L1 norm in the image space, and
\begin{equation}
J_\mathcal{Y}^\textrm{rec} = \operatorname{MCE}(\boldsymbol{y}^\textrm{rec}, \bar{\boldsymbol{y}}) = -\frac{1}{H W} \sum_{i \in \mathcal{I}} \bar{\boldsymbol{y}}_i^\textrm{T} \log \boldsymbol{y}_i^\textrm{rec}
\end{equation}
in the segmentation space, respectively, both comparing the ground truth input with the reconstruction and by this optimizing cycle consistency.
The losses can be grouped into the forward cycle loss
\begin{equation}
\label{eq:forward}
J^\textrm{forward} = \alpha \cdot J^\textrm{seg} + (1 - \alpha) \cdot J_\mathcal{X}^\textrm{rec}
\end{equation}
and the backward cycle loss
\begin{equation}
\label{eq:backward}
J^\textrm{backward} = \beta \cdot J^\textrm{adv} + (1 - \beta) \cdot J_\mathcal{Y}^\textrm{rec},
\end{equation}
with the factors $0 \le \alpha, \beta, \gamma \le 1$.
Weighted addition of forward and backward loss then yields the total generator loss
\begin{equation}
\label{eq:gen_loss}
J^\textrm{gen} = \gamma \cdot J^\textrm{forward} + (1 - \gamma) \cdot J^\textrm{backward}.
\end{equation}

The \textit{discriminator training} (see lower part of Fig.~\ref{fig:cycle_gan}) takes place in the same iteration, uses the same input data as the generators, and is split into two loss paths:
The first path evaluates a real input image $\bar{\boldsymbol{x}}$ and generates the estimate $\boldsymbol{P}^\textrm{real}_\mathcal{X}$.
In the second path, $\mathcal{F}$ generates the synthetic image $\boldsymbol{x}$ and inserts it into a ringbuffer~\cite{Shrivastava2017}.
Sampling from this buffer yields the generated image $\boldsymbol{x}^\prime$ (can originate from an earlier training step), which is processed by the discriminator to generate the estimate $\boldsymbol{P}^{\textrm{gen} \prime}_\mathcal{X}$.
The ringbuffer prevents forgetting during the discriminator learning~\cite{Shrivastava2017}.
In each iteration, the discriminator of the least squares GAN (LS-GAN)~\cite{Mao2016} is trained by the discriminator loss
\begin{equation}
\label{eq:disc_loss}
J^\textrm{disc} = \operatorname{MSE}(\boldsymbol{P}^\textrm{real}_\mathcal{X}, \boldsymbol{1}) + \operatorname{MSE}(\boldsymbol{P}^{\textrm{gen} \prime}_\mathcal{X}, \boldsymbol{0}),
\end{equation}
with $\boldsymbol{1}$ and $\boldsymbol{0}$ being vectors of ones and zeros, respectively.

\textbf{Training Protocol.}
The ordered steps for learning the entire framework, \ie, the iterative batch training protocol, are summarized in Table~\ref{tab:protocol}.
In general, the protocol can be categorized into executing the networks and functions (steps 1 and 2), generator learning (steps 3 and 4), and discriminator learning (steps 5 and 6).

% Training Protocol
\begin{table}[t!]
	\centering
	\setlength{\tabcolsep}{4pt}
	\begin{tabularx}{\columnwidth}{rX}
		\toprule
		1. & Load \textbf{inputs} (image $\bar{\boldsymbol{x}}$ and respective segmentation $\bar{\boldsymbol{y}}$)                                 \\
		2. & Compute \textbf{outputs} of all networks and functions ($\mathcal{G}, \mathcal{F}, \mathcal{D}_\mathcal{X}$, noise injection, softmax) \\ \midrule
		3. & Compute \textbf{generator loss} $J^\textrm{gen}$ \eqref{eq:gen_loss}                                                                   \\
		4. & Backpropagate gradients and \textbf{update weights} of $\mathcal{G}$ and $\mathcal{F}$ ($\mathcal{D}_\mathcal{X}$ weights fixed)       \\ \midrule
		5. & Compute \textbf{discriminator loss} $J^\textrm{disc}$ \eqref{eq:disc_loss}                                                             \\
		6. & Backpropagate gradients and \textbf{update weights} of $\mathcal{D}_\mathcal{X}$ ($\mathcal{G}$ and $\mathcal{F}$ weights fixed)       \\ \bottomrule
	\end{tabularx}
	\caption{
		\textbf{Training protocol} for a batch of data (one iteration).
		The generators are updated (steps 3 and 4) before the discriminator (steps 5 and 6).
		While updating the generator, the weights of the discriminator are fixed and vice versa.
	}
	\label{tab:protocol}
%	\vspace*{-0.25cm}
\end{table}

\subsection{Noise Injection}
\label{sec:methodology_noise}
To effectively suppress watermarks in the forward cycle, we use a noise injection (block in Fig.~\ref{fig:cycle_gan}) in the image reconstruction step of the forward cycle.
Since the noise basically can be of any kind, we perform experiments with a so-called maximum-only noise, quantization noise (naturally correlated), and uncorrelated additive Gaussian noise.

% Argmax Noise
\textbf{Maximum-only Noise.}
The maximum-only noise is applied as noise injection baseline used with the CycleGAN architecture.
Setting for each pixel all non-maximum class components in the segmentation prediction $\tilde{\boldsymbol{y}}$ simply to zero, we obtain the noisy segmentation prediction $\hat{\boldsymbol{y}}$, introducing an error into the signal, which is measurable as noise.

% Quantization
\textbf{Quantization Noise.}
Quantization means amplitude discretization~\cite{Gray1984}.
A common use case of quantization is the reduction of the amount of bits required for data transmission (bitrate)~\cite{gersho2012, Loehdefink2020IV} or storage (file size), \eg, in a file system~\cite{Rippel2017, Theis2017, rippel2018learned}.
It reduces information and therefore also leads to an error, the so-called quantization noise.

The quantization noise originates from an $n$-bit scalar uniform quantizer, using a codebook $\mathit{CB} = \{-1, ..., 1\}$, with $|\mathit{CB}| = 2^n$ uniformly distributed reconstruction levels.
The reconstruction level with the minimal distance to the scalar input value $\tilde{y}_{i,s}$, \ie, class $s \in \mathcal{S}$ of pixel $i\in\mathcal{I}$ from the segmentation prediction $\tilde{\boldsymbol{y}}$, is the scalar output of the quantizer (MSE criterion)
\begin{equation}
	\label{eq:quant_noise}
	\hat{y}_{i,s} = \underset{c \in \mathit{CB}}{\arg\min}\left(\tilde{y}_{i,s} - c\right)^2,
\end{equation}
obtaining class $s \in \mathcal{S}$ of pixel $i\in\mathcal{I}$ in the noisy segmentation prediction $\hat{\boldsymbol{y}}$.
While any approximation function can be used for backpropagation~\cite{bengio-arxiv13-condcomp, loehdefink20ITSC}, in this work, we simply approximate the quantizer by the identity function to preserve the gradients $\nabla \tilde{\boldsymbol{y}} = \nabla \hat{\boldsymbol{y}}$.

% Gaussian Noise
\textbf{Gaussian Noise.}
The last type of noise stems from a noise generator providing additive Gaussian noise, which is uncorrelated to the input signal of the adder.
By sampling from the normal distribution $\mathcal{N}$ with a mean of $\tilde{y}_{i,s}$, and the standard deviation $\sigma$, we obtain the noisy segmentation prediction elements
\begin{equation}
	\label{eq:gauss_noise}
	\hat{y}_{i,s} \sim \mathcal{N}(\tilde{y}_{i,s}, \sigma^2),
\end{equation}
again with pixel $i\in\mathcal{I}$ and class $s\in\mathcal{S}$.
So in contrast to the quantization noise, the Gaussian noise is random.

\section{Evaluation}
\label{sec:eval}
In this section, we first explain the metrics that will be used to evaluate the proposed methods.
Then we present the experimental setup used in training and inference.
Afterwards, we will visualize the effects of the proposed noise injection in the segmentation space, investigate the segmentation output by means of the intersection over union (IoU), and show the mIoU and PSNR performances of all experiments performed.

\subsection{Metrics}
The following metrics are used in this work.

% SNR
\textbf{Signal-to-Noise Ratio.}
To quantify the distortion in the forward cycle and compare the two noise types, we compute the signal-to-noise ratio
\begin{align}
\label{eq:snr_qnoise}
\mathit{SNR} & = 10 \log\left(\frac{\sum_{i\in \mathcal{I}}\norm{\tilde{\boldsymbol{y}}_i}_2^2}{\sum_{i\in \mathcal{I}}\norm{\hat{\boldsymbol{y}}_i - \tilde{\boldsymbol{y}}_i}_2^2}\right) [\si{\decibel}] \\
\label{eq:snr_gnoise}
& = 10 \log\left(\frac{\sum_{i\in \mathcal{I}}\norm{\tilde{\boldsymbol{y}}_i}_2^2}{|\mathcal{I}|\cdot \sigma^2}\right) [\si{\decibel}]
\end{align}
by measuring the noisy segmentation predictions $\hat{\boldsymbol{y}}$ and the segmentation predictions $\tilde{\boldsymbol{y}}$ for the quantization noise \eqref{eq:snr_qnoise}.
When using Gaussian noise, the SNR can alternatively be computed via the noise variance $\sigma^2$ \eqref{eq:snr_gnoise}.
Consequently, to set the SNR for the Gaussian noise adder, the image-dependent standard deviation
\begin{equation}
\label{eq:sigma}
\sigma(\mathit{SNR}, \tilde{\boldsymbol{y}}) = \tfrac{1}{\sqrt{|\mathcal{I}|}} 10^{\frac{1}{20}(10\log(\sum_{i\in \mathcal{I}}\norm{\tilde{\boldsymbol{y}}_i}_2^2)-\mathit{SNR})},
\end{equation}
can be specified, depending on the segmentation prediction $\tilde{\boldsymbol{y}}$.
In our experiments, we measure the SNR values obtained from the quantizers and set the exact same SNRs for the Gaussian noise adder to achieve better comparability.
Note that the SNR in case of the quantization noise injection is an averaged result from the validation set, while for the Gaussian noise injection, every single image has the desired SNR, since $\sigma$ is specified.

% PSNR
\textbf{Peak-Signal-to-Noise Ratio.}
To evaluate the image reconstruction performance in the forward cycle, we compute the peak-signal-to-noise ratio~\cite{Salomon2004}
\begin{equation}
\label{eq:psnr}
\mathit{PSNR} = 10 \log\left(\frac{\bar{x}_\textrm{max}^2}{\frac{1}{C \cdot H \cdot W}\sum_{i\in\mathcal{I}} \norm[\big]{ \bar{\boldsymbol{x}}_i-\hat{\boldsymbol{x}}^\textrm{rec}_i}^2_2}\right) [\si{\decibel}]
\end{equation}
between the ground truth image $\bar{\boldsymbol{x}}$ and the reconstruction $\hat{\boldsymbol{x}}^\textrm{rec}$ with the maximum pixel fluctuation ${\bar{x}_\textrm{max}} = 2$.

% Visualization
\begin{figure*}[t!]
	\centering
	\subfloat{
		\includegraphics[width=0.19\textwidth]{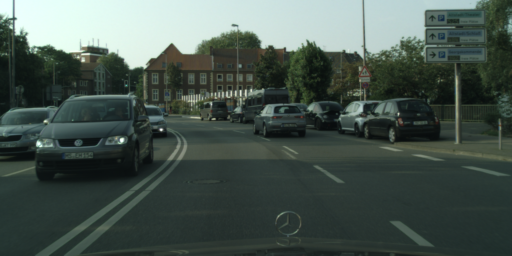}
	}
	\subfloat{
		\stackunder[5pt]{\includegraphics[width=0.19\textwidth]{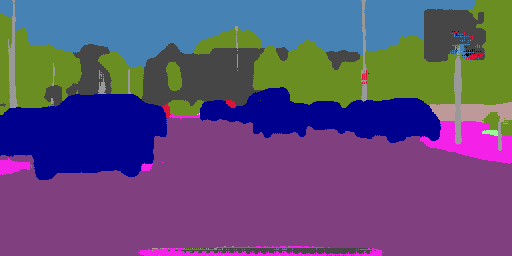}}{$\mathit{mIoU} = \SI{56.2}{\percent}$}
	}
	\subfloat{
	\stackunder[5pt]{\includegraphics[width=0.19\textwidth]{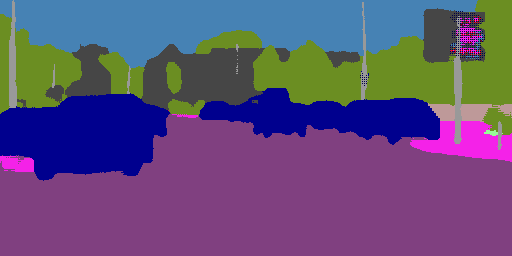}}{$\mathit{mIoU} = \SI{45.3}{\percent}$}
	}
	\subfloat{
		\stackunder[5pt]{\includegraphics[width=0.19\textwidth]{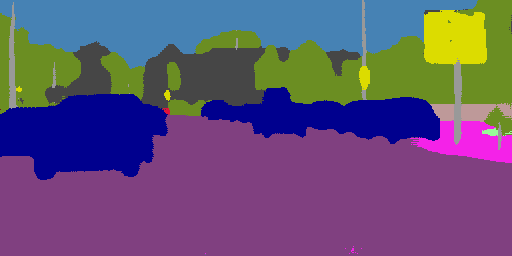}}{$\mathit{mIoU} = \SI{67.9}{\percent}$}
	}
	\subfloat{
		\includegraphics[width=0.19\textwidth]{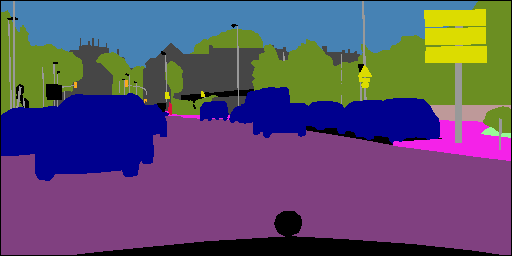}
	}
	\\
	\subfloat{
		\includegraphics[width=0.19\textwidth]{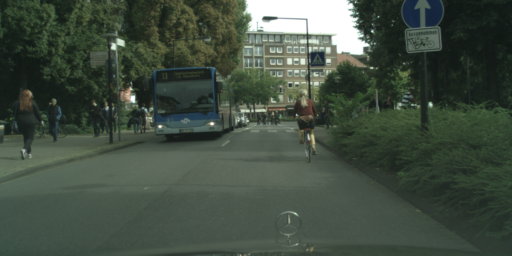}
	}
	\subfloat{
		\stackunder[5pt]{\includegraphics[width=0.19\textwidth]{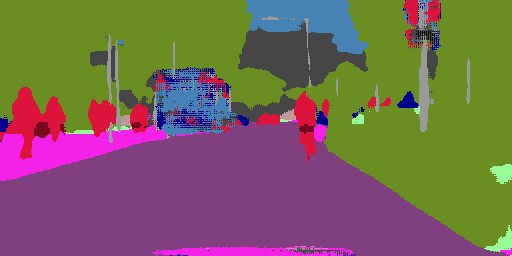}}{$\mathit{mIoU} = \SI{30.2}{\percent}$}
	}
	\subfloat{
	\stackunder[5pt]{\includegraphics[width=0.19\textwidth]{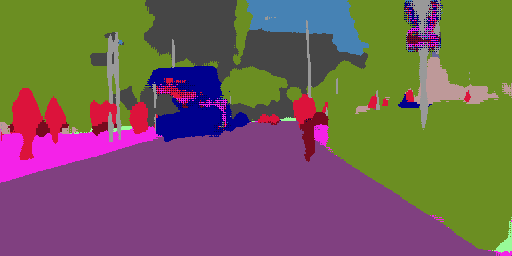}}{$\mathit{mIoU} = \SI{29.9}{\percent}$}
	}
	\subfloat{
		\stackunder[5pt]{\includegraphics[width=0.19\textwidth]{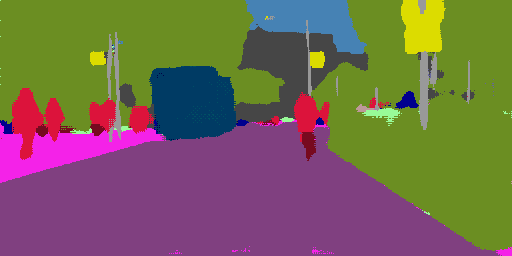}}{$\mathit{mIoU} = \SI{40.6}{\percent}$}
	}
	\subfloat{
		\includegraphics[width=0.19\textwidth]{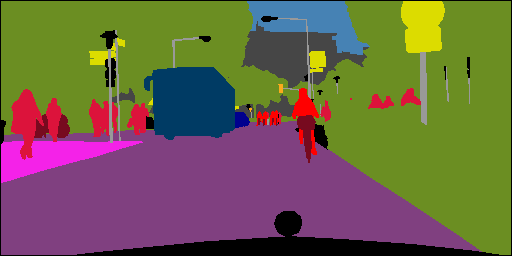}
	}
	\\
	\setcounter{subfigure}{0}
	\subfloat[Input image $\bar{\boldsymbol{x}}$\label{fig:samples_input}]{
		\stackunder[5pt]{\includegraphics[width=0.19\textwidth]{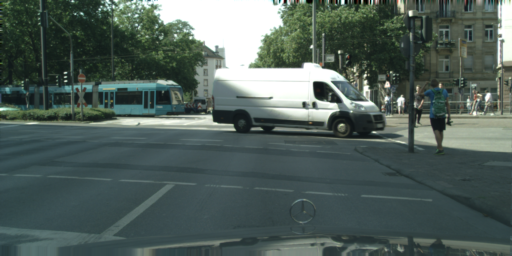}}{\phantom{$\mathit{mIoU} = \SI{0}{\percent}$}}
	}
	\subfloat[Pred. segmentation mask, ERFNet\label{fig:samples_erfnet}]{
		\stackunder[5pt]{\includegraphics[width=0.19\textwidth]{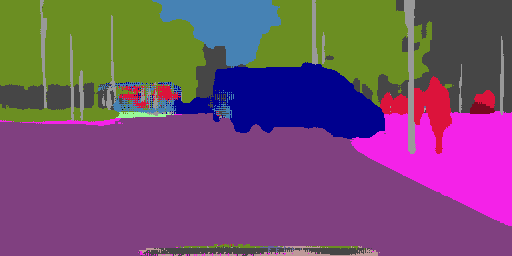}}{$\mathit{mIoU} = \SI{39.4}{\percent}$}
	}
	\subfloat[Pred. segmentation mask, CycleGAN ERFNet \textbf{without} quantization noise injection in training\label{fig:samples_no_noise}]{
	\stackunder[5pt]{\includegraphics[width=0.19\textwidth]{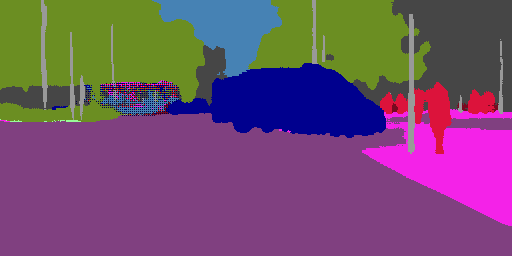}}{$\mathit{mIoU} = \SI{38.4}{\percent}$}
	}
	\subfloat[Pred. segmentation mask, CycleGAN ERFNet \textbf{with} quantization noise injection in training\label{fig:samples_q_noise_2_bit}]{
		\stackunder[5pt]{\includegraphics[width=0.19\textwidth]{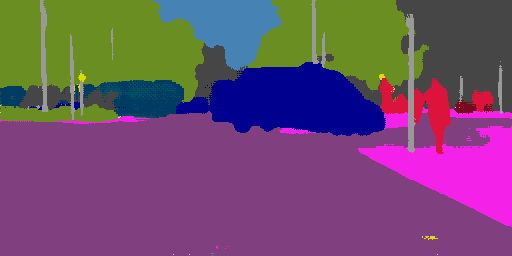}}{$\mathit{mIoU} = \SI{41.3}{\percent}$}
	}
	\subfloat[Ground truth $\bar{\boldsymbol{y}}$\label{fig:samples_gt}]{
		\stackunder[5pt]{\includegraphics[width=0.19\textwidth]{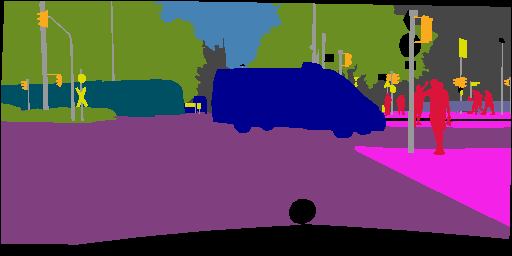}}{\phantom{$\mathit{mIoU} = \SI{0}{\percent}$}}
	}
	\caption{
		\textbf{Segmentation samples from the ERFNet baseline and the CycleGAN framework using also ERFNet, with our proposed quantization noise injection} originating from CS$^\textrm{val}$~\cite{Cordts2016}.
		Based on \protect\subref{fig:samples_input} the input image, the comparison shows \protect\subref{fig:samples_erfnet} the predicted segmentation masks of the ERFNet~\cite{Romera2018}, \protect\subref{fig:samples_no_noise} the CycleGAN ERFNet without noise injection, and \protect\subref{fig:samples_q_noise_2_bit} the CycleGAN ERFNet with 2-bit quantization noise injection along with \protect\subref{fig:samples_gt} the ground truth $\bar{\boldsymbol{y}}$.
		The quantization noise injection in the forward cycle enables the CycleGAN ERFNet to recognize the classes ``traffic sign'', ``bus'', and ``train''.
	}
	\label{fig:samples}
	\vspace*{-0.25cm}
\end{figure*}

% mIoU
\textbf{Intersection Over Union.}
The intersection over union (IoU) and the mean intersection over union (mIoU) are commonly used quality measures for semantic segmentation.
For the class with index $s \in \mathcal{S}$, we compute the intersection over union
\begin{equation}
\label{eq:iou}
\mathit{IoU}_s = \frac{\mathit{TP_s}}{\mathit{TP_s} + \mathit{FP_s} + \mathit{FN_s}},
\end{equation}
with the true positives $\mathit{TP_s}$, the false positives $\mathit{FP_s}$, and the false negatives $\mathit{FN_s}$.
The average over all classes then delivers the mean intersection over union
\begin{equation}
\label{eq:miou}
\mathit{mIoU} = \frac{1}{|\mathcal{S}|} \sum_{s\in \mathcal{S}}\mathit{IoU}_s.
\end{equation}
Both intersection metrics are used for the evaluation of the ERFNet~\cite{Romera2018} and the CycleGAN~\cite{mondal2019revisiting} with the proposed noise injection.

\subsection{Experimental Setup}
\label{sec:eval_setup}
This section addresses the data set, generator and discriminator architectures, as well as training parameters used in our experiments.

\textbf{Data.}
% Table Cityscapes
We use the Cityscapes (CS) dataset~\cite{Cordts2016}, consisting of 2,975 annotated training images (CS$^\textrm{train}$), 500 annotated validation images  (CS$^\textrm{val}$), and 1,525 test images (CS$^\textrm{test}$) without ground truth labels.
Although the native resolution of the dataset is $2048 \times 1024$, we use a downsampled resolution of $512 \times 256$ for our experiments since otherwise the available GPU memory is exceeded.

\textbf{Generator.}
% Gen-Architectures
For comparability with a state-of-the-art semantic segmentation, the generators used in the CycleGAN architecture also are based on the ERFNet~\cite{Romera2018}.
The input and output dimensions for the two generators in our framework are different, since the image space has $C=3$ channels, while the segmentation space has $S=20$ channels---one for each class, including background.
So for the generator $\mathcal{G}$, we use the original ERFNet without modifications.
However, for the generator $\mathcal{F}$, we adjust the input to consist of $S$ channels.
Due to the parallel convolutional and pooling-based downsampling in the ERFNet architecture, the number of feature maps in the encoder must always increase.
So instead of the original implementation with 16 feature maps after the first layer, we use 32 feature maps for generator $\mathcal{F}$.
The output of $\mathcal{F}$ consists of $C$ feature maps representing the color channels in the image space.

\textbf{Discriminator.}
Concerning the details of the discriminator architecture, we refer the interested reader to the \mbox{CycleGAN} work by Zhou \etal~\cite{Zhu2017}, since we adopted the implementation of a least squares patchGAN discriminator with instance normalization.
For the purpose of dimensionality reduction, in this architecture the output consists of height $H_\textrm{D} = (\frac{H}{8}-2)$ and width $W_\textrm{D} = (\frac{W}{8}-2)$ with $H$ and $W$ being height and width of the input image, respectively, resulting from the three convolutional layers with a stride of 2.
As explained in Section \ref{sec:methodology_architecture}, we update the discriminator by a history of generated images~\cite{Shrivastava2017}, stored in a ringbuffer.

% Class-Wise IoU
\begin{figure*}[t!]
	\centering
	\includegraphics{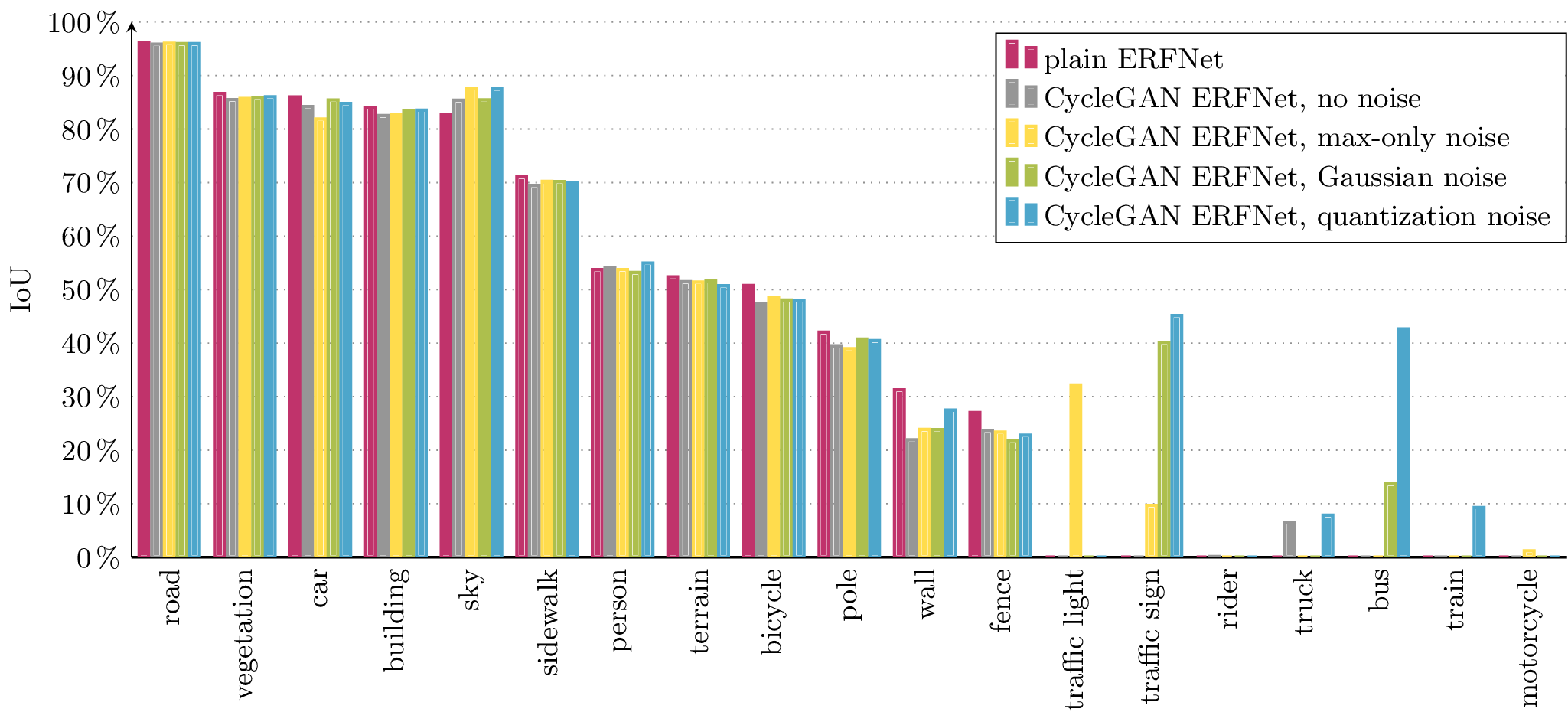}
	\caption{
		\textbf{Class-wise evaluation of the intersection over union (IoU) \eqref{eq:iou}} on the Cityscapes validation set~\cite{Cordts2016}, with $S=20$ classes (without background).
		The proposed noise injection with max-only, quantization, or additive Gaussian noise is compared to the ERFNet and the CycleGAN ERFNet without noise injection.
		The SNR is \SI{0.22}{\decibel} for the max-only noise injection, \SI{7.38}{\decibel} for the 2-bit quantization noise injection, and \SI{21.02}{\decibel} for the Gaussian noise injection.
		Figure is best seen in color.
	}
	\label{fig:plot_iou}
%	\vspace*{-0.25cm}
\end{figure*}

\textbf{Training Parameters.}
The framework is trained for 200 epochs~\cite{Zhu2017} in total with a batch size of 4, an initial learning rate of 0.0002, and the ADAM optimizer~\cite{Kingma2015}.
The first 100 epochs use the initial learning rate as basis for the ADAM optimizer, afterwards, the learning rate linearly decays to zero at epoch 200~\cite{Zhu2017}.
The ADAM optimizer uses the parameters $\beta_1 = 0.5$ and $\beta_2 = 0.999$.
The weights of the networks are initialized with a normal distribution with zero mean and a standard deviation of 0.02.
The hyperparameters for the loss functions defined in \eqref{eq:forward}, \eqref{eq:backward}, and \eqref{eq:gen_loss} are $\alpha = 0.5$, $\beta = \frac{1}{11}$, and $\gamma = \frac{20}{31}$.
The ringbuffer, used to store generated images $\boldsymbol{x}$ for the discriminator training, has a size of 50 images~\cite{Zhu2017}, and read access is organized in a random fashion.

\subsection{Visual Effects of Noise Injection}
\label{sec:visualization}
% Description
From left to right, Figure~\ref{fig:samples} shows an input image $\bar{\boldsymbol{x}}$ from the Cityscapes validation set with the respective predicted segmentation masks showing the most probable class from prediction $\boldsymbol{y}$.
Results are shown for the plain ERFNet and the CycleGAN ERFNet without and with the proposed injection of 2-bit quantization noise along with the ground truth segmentation mask on the right side.

% New Objects Recognized
While the overall segmentation performance seems to be quite similar for both of the approaches at first sight, \ie, both networks generate reasonable segmentation masks without major flaws, it is noteworthy that additional objects appear when using the CycleGAN ERFNet with noise injection in training (column \protect\subref{fig:samples_q_noise_2_bit}) instead of the ERFNet (column \protect\subref{fig:samples_erfnet}) or the CycleGAN ERFNet without noise injection (column \protect\subref{fig:samples_no_noise}).
In the presented example images this concerns the classes ``traffic sign'' (in each example), ``bus'' (second row), and ``train'' (third row), not being recognized by the ERFNet baseline but being rather well segmented by the CycleGAN ERFNet with quantization noise injection.
This also affects the mIoU, causing large improvements by, \eg, \SI{11.7}{\percent} absolute in the first row of Figure~\ref{fig:samples}.

\subsection{Class-Wise Intersection Over Union}
\label{sec:val_iou}
% Description
Figure~\ref{fig:plot_iou} shows the intersection over union (IoU) \eqref{eq:iou} for each individual class of Cityscapes~\cite{Cordts2016} as a bar chart for the validation dataset CS$^\textrm{val}$.
For each of the $S=20$ classes (except background), the performance of the ERFNet, the CycleGAN ERFNet without noise injection in training, the CycleGAN ERFNet with injection of max-only noise, Gaussian noise, and quantization noise are compared.

% Approximately En Par
The majority of classes is predicted as expected, where all the approaches are approximately en par, with the ERFNet in most cases being only slightly better.
% Some Classes Not Predicted
Sur\-pri\-sing\-ly, there are several classes in the chart having an IoU of zero, for one or more segmentation approaches.
\mbox{Reporting} the results on CS$^\textrm{val}$, this means that there are classes, which are never predicted by these specific approaches in this dataset.
In total, there are three classes that are still not predicted by any approach.
However, we find the four classes ``traffic sign'', ``truck'', ``bus'', and ``train'', which are not detected by the ERFNet baseline, but by our CycleGAN ERFNet with the proposed 2-bit quantization noise injection.
% Resolution is the Reason
Not detecting some of the classes suggests that the ERFNet, which in the original work of Romera \etal~\cite{Romera2018} actually achieves an mIoU of \SI{71.5}{\percent} on the Cityscapes validation set, may be a questionable segmentation baseline in this case.
We assume, the cause for this shortcoming is the reduced image resolution of $256 \times 512$ compared to the original resolution of $512 \times 1024$~\cite{Romera2018}, which was necessary to cope with the 11 GB video memory of the \texttt{NVIDIA GeForce 1080Ti} GPUs during training.
Nevertheless, our proposed noise injection is able to partly cure the problem even in the smaller image resolution, which we deem to be an important result.

% Results Table
\begin{table}[t!]
	\centering
	\setlength{\tabcolsep}{2.5pt}
	\begin{tabular}{crrrr}
		\toprule
		        \textbf{Network Type,}          &     \multicolumn{2}{c}{\textbf{Noise}}      &         \textbf{mIoU} &          \textbf{PSNR} \\
		          \textbf{Noise Type}           & \textbf{\# of Bits} &     \textbf{SNR [dB]} &         \textbf{[\%]} &          \textbf{[dB]} \\ \midrule
		             plain ERFNet,              & \multirow{2}{*}{--} &   \multirow{2}{*}{--} & \multirow{2}{*}{40.2} &    \multirow{2}{*}{--} \\
		               no noise                 &                     &                       &                       &                        \\ \midrule
		           CycleGAN ERFNet,             & \multirow{2}{*}{--} &   \multirow{2}{*}{--} & \multirow{2}{*}{39.4} & \multirow{2}{*}{25.37} \\
		               no noise                 &                     &                       &                       &                        \\ \midrule
		           CycleGAN ERFNet,             & \multirow{2}{*}{--} & \multirow{2}{*}{0.22} & \multirow{2}{*}{41.4} & \multirow{2}{*}{24.71} \\
		               max-only                 &                     &                       &                       &                        \\ \midrule
		                                        &                  -- &                 40.00 &                  38.7 &                  24.79 \\
		                                        &                  -- &                 30.00 &                  39.0 &                  25.05 \\
		                                        &                  -- &                 21.02 &                  42.1 &                  24.96 \\
		                                        &                  -- &                 14.50 &                  41.5 &                  24.43 \\
		\multirow{2}{*}[-2pt]{CycleGAN ERFNet,} &                  -- &                  7.38 &                  38.9 &                  23.44 \\
		    \multirow{2}{*}[-2pt]{Gaussian}     &                  -- &                 -0.55 &                  39.6 &                  22.82 \\
		                                        &                  -- &                -10.00 &                  41.9 &                  20.77 \\
		                                        &                  -- &                -15.00 &                  39.3 &                  19.56 \\
		                                        &                  -- &                -20.00 &                  39.5 &                  18.74 \\ \midrule
		                                        &                   4 &                 21.02 &                  43.3 &         \textbf{25.54} \\
		           CycleGAN ERFNet,             &                   3 &                 14.50 &                  42.9 &                  24.48 \\
		             quantization               &                   2 &                  7.38 &         \textbf{45.1} &                  25.30 \\
		                                        &                   1 &                 -0.55 &                   7.7 &                  24.38 \\ \bottomrule
	\end{tabular}
	\caption{
		\textbf{Segmentation and image reconstruction results}, measured by mIoU~\eqref{eq:miou} and PSNR~\eqref{eq:psnr}, respectively, on the Cityscapes validation set CS$^\textrm{val}$.
		% CycleGAN
		As baselines, we use a traditional semantic segmentation training setting, where $J^\textrm{seg}$ is the only loss function (plain ERFNet), and our cycle-consistent architecture (CycleGAN ERFNet).
		% SNR
		We then compare injection of ``max-only'', Gaussian \eqref{eq:gauss_noise}, and quantization \eqref{eq:quant_noise} noise (partly) with identical levels of signal-to-noise ratio (SNR) in the latent segmentation space.
		The SNR values were measured for the quantization noise injection~\eqref{eq:snr_qnoise} and afterwards transferred to the Gaussian noise adder to obtain comparable results.
		Best results in \textbf{boldface}.
	}
	\label{tab:val_finely_annotated}
%	\vspace*{-0.25cm}
\end{table}

\subsection{Mean Intersection Over Union and PSNR}
\label{sec:eval_noise}
% Intro
With the ERFNet as segmentation baseline, Table~\ref{tab:val_finely_annotated} compares various noise types for injection into the forward cycle of the CycleGAN architecture, namely no noise, max-only noise, Gaussian noise, and quantization noise, see Section~\ref{sec:methodology_noise}.
It shows the mIoU \eqref{eq:miou} for all experiments and additionally the PSNR \eqref{eq:psnr}, when the CycleGAN ERFNet is used.
For experiments with noise injection, also the signal-to-noise ratio (SNR) \eqref{eq:snr_qnoise} is shown, being measured in the cases of the max-only and quantization noise and adjusted for the Gaussian noise by \eqref{eq:sigma}.
For the Gaussian noise injection, we chose the SNR to match the SNR of the quantization noise injection with $n = \{1,2,3,4\}$ bit, beside additional experiments with SNRs of \SI{-20}{\decibel}, \SI{-15}{\decibel}, \SI{-10}{\decibel}, \SI{30}{\decibel}, and \SI{40}{\decibel}.

% mIoU and PSNR
Comparing the ERFNet to the CycleGAN ERFNet without noise injection, an mIoU drop of \SI{0.8}{\percent} absolute can be observed, applying the max-only noise, however, obtains an mIoU increase of \SI{1.2}{\percent} absolute compared to the baseline.
The best mIoU when injecting Gaussian noise is \SI{42.1}{\percent}, which is an improvement by \SI{1.9}{\percent} absolute over the ERFNet baseline.
The overall best approach in all our experiments though is the 2-bit quantization noise, corresponding to an SNR of \SI{7.38}{\decibel}, with an mIoU of \SI{45.1}{\percent}, significantly improving the CycleGAN ERFNet without noise injection by \SI{5.7}{\percent} absolute and the plain ERFNet by \SI{4.9}{\percent} absolute.
For most of the experiments, the PSNR fluctuates only slightly, where the 4-bit quantization noise with \SI{25.54}{\decibel} has the highest PSNR and the CycleGAN ERFNet without noise injection having the second best PSNR of \SI{25.37}{\decibel}, which could have been expected.
For the Gaussian noise injection, a very low SNR leads to a degraded image reconstruction performance with a PSNR of only \SI{18.74}{\decibel} at an SNR of \SI{-20}{\decibel}, while for the quantization noise injection, the lowest SNR of \SI{-0.55}{\decibel} results in a collapsed mIoU of \SI{7.7}{\percent}.

% Interpretations 
Since the noise injection into the latent segmentation space increases the performance of the CycleGAN compared to the CycleGAN without noise injection, we assume watermarks to be effectively avoided, and therefore the training process of the generators $\mathcal{G}$ and $\mathcal{F}$ to be optimized.
Interestingly, the obtained PSNR of \SI{25.30}{\decibel} with 2-bit quantization noise is very close to the PSNR of \SI{25.37}{\decibel} without noise injection.
The visual examples from Section~\ref{sec:visualization} show that the CycleGAN ERFNet with noise injection detects more classes than the ERFNet baseline, including ``traffic signs'', obviously leading to an increased mIoU.
Furthermore, Table~\ref{tab:val_finely_annotated} shows that the proposed noise injection is more effective with the correlated quantization noise than with the uncorrelated Gaussian noise, why we conclude that the information bottleneck enforced by the quantization noise is an important aspect for avoiding watermarks in the segmentation space, and thereby obtain powerful networks that have been learned ``under stress''.

\section{Conclusion}
\label{sec:conclusion}
In this paper, we proposed a noise injection into the latent segmentation space of a cycle-consistent semantic segmentation framework to overcome the problem of watermark propagation in the image $\rightarrow$ segmentation $\rightarrow$ image cycle.
The proposed noise injection was able to avoid the watermarks and improve the training process of the generator networks, used for segmentation and image reconstruction, so that in comparison to the baseline more classes of the Cityscapes dataset could be recognized, including the highly relevant class ``traffic sign'', which was also shown in visual examples originating from the framework.
By using a 2-bit quantization noise injection, an absolute mIoU improvement of \SI{4.9}{\percent} over the ERFNet baseline was achieved.
We found that quantization noise is better suited to increase the segmentation performance in the investigated framework than additive random Gaussian noise.

\ifwacvfinal
\textbf{Acknowledgement.}
The authors gratefully acknowledge support of this work by Daniel Wilke.
\fi

%\vfill

{
	\small
	\bibliographystyle{ieee}
	\bibliography{ifn_spaml_bibliography}
}

\end{document}